\documentclass{article} 
\usepackage{iclr2021_conference,times}

\usepackage{amsmath,amsfonts,bm}

\def\eqref#1{equation~\ref{#1}}

\def\1{\bm{1}}

\DeclareMathAlphabet{\mathsfit}{\encodingdefault}{\sfdefault}{m}{sl}
\SetMathAlphabet{\mathsfit}{bold}{\encodingdefault}{\sfdefault}{bx}{n}

\usepackage{booktabs}

\usepackage{hyperref}
\usepackage{url}
\usepackage{graphicx} 
\usepackage{caption}
\usepackage{tikz}
\usepackage{amsmath, amssymb}
\usepackage{algorithm}
\usepackage{algpseudocode}
\usepackage{graphicx}
\usepackage{hyperref}
\usepackage{wrapfig}
\usepackage{graphicx}
\usepackage{caption}
\usepackage{booktabs}
\usepackage[table]{xcolor}

\usepackage{booktabs}
\usepackage[table]{xcolor}
\usepackage{tabularx}
\usepackage[table]{xcolor}
\usepackage{multirow}
\usepackage{array}

\usepackage{booktabs}
\usepackage[table]{xcolor}
\usepackage{tabularx} 
\usetikzlibrary{shapes, arrows, positioning, fit}

\iclrfinalcopy

\title{MetaWorld: Skill Transfer and Composition in a Hierarchical World Model for Grounding High-Level Instructions}

\author{
  \textbf{Yutong Shen}$^1$, \textbf{Hangxu Liu}$^2$, \textbf{Kailin Pei}$^1$, \textbf{Ruizhe Xia}$^1$, \textbf{Tongtong Feng}$^{3, \ast}$ \\
  $^1$Beijing University of Technology \quad $^2$Fudan University \quad $^3$Tsinghua University \\
  \texttt{syt2004@emails.bjut.edu.cn}
}
\begin{document}

\maketitle

\begin{figure*}[h] 
    \centering 
    \includegraphics[width=\textwidth]{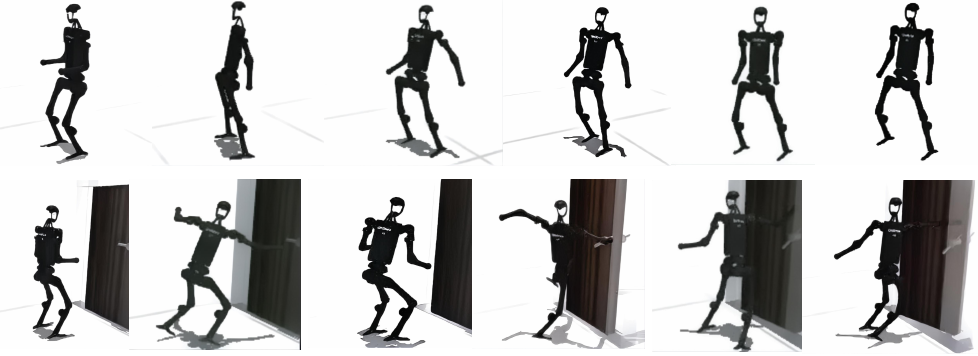} 
    \caption{This figure demonstrates the effectiveness of our hierarchical world model architecture on both the walking expert policy and the door-opening task.} 
    \label{fig:zhanshi} 
\end{figure*}

\begin{abstract}

Humanoid robot loco-manipulation remains constrained by the semantic-physical gap. Current methods face three limitations: Low sample efficiency in reinforcement learning, poor generalization in imitation learning, and physical inconsistency in VLMs. We propose \textbf{MetaWorld}, a hierarchical world model that integrates semantic planning and physical control via expert policy transfer. The framework decouples tasks into a VLM-driven semantic layer and a latent dynamics model operating in a compact state space. Our dynamic expert selection and motion prior fusion mechanism leverages a pre-trained multi-expert policy library as transferable knowledge, enabling efficient online adaptation via a two-stage framework. VLMs serve as semantic interfaces, mapping instructions to executable skills and bypassing symbol grounding. Experiments on Humanoid-Bench show \textbf{MetaWorld} outperforms world model-based RL in task completion and motion coherence. Our code will be found at \url{https://anonymous.4open.science/r/metaworld-2BF4/}
\end{abstract}

\section{Introduction}

Humanoid robots, as quintessential embodiments of embodied intelligence, have long faced a fundamental challenge: executing semantically-driven loco-manipulation tasks within unstructured and dynamic environments. At the core of this challenge lies a significant ``abstraction gap'' in current robot control systems---the disconnect between high-level semantic understanding and low-level physical execution~\cite{kunze2011towards,huang2024rekep,geng2024sage,zhang2024universal}. On one hand, large-scale models represented by Vision-Language Models (VLMs) demonstrate exceptional capabilities in high-level task planning and semantic reasoning, understanding ``what to do''; on the other hand, low-level control methods based on imitation learning or reinforcement learning can generate precise joint-level actions, addressing ``how to do it.'' However, the absence of a unified and scalable framework to bridge these disparate capabilities often results in semantic plans that are decoupled from physical constraints, conversely, low-level control policies often lack the versatility to generalize across complex, composed high-level tasks.

Current mainstream approaches exhibit distinct limitations in addressing this problem. Although end-to-end reinforcement learning is theoretically capable of discovering optimal policies, it is often prohibitively sample-inefficient, rendering it impractical for real-world robotic platforms~\cite{song2024hagrasp}. Imitation learning, while effective at acquiring natural motion patterns from human demonstrations, typically yields policies with limited robustness that remain highly sensitive to environmental fluctuations and external disturbances~\cite{lesort2024evil}. Direct application of VLMs to robot control faces severe ``symbol grounding'' problems, with generated action plans often being kinematically or dynamically infeasible. While existing literature explores the integration of these methods, most approaches remain restricted to simplistic task concatenation or a loose coupling of independent modules, failing to establish a truly unified, hierarchical interaction architecture~\cite{pan2025omnimanip,li2025hamster,shen2025detachcrossdomainlearninglonghorizon}.

To systematically address these challenges, this paper presents \textbf{MetaWorld}, a hierarchical world model-based robot control framework. The proposed framework synergistically combines the semantic reasoning of VLMs, motion priors from imitation learning, and online adaptation mechanisms from model-based reinforcement learning, establishing a cohesive and scalable pipeline that bridges the gap between high-level semantic reasoning and low-level physical execution.
\textbf{The main contributions of this paper include the following three aspects}:

\begin{itemize}
    \item We propose a hierarchical, modular world model architecture that enables multi-scale decomposition of task representations.
    \item We introduce a dynamic expert selection and motion prior fusion mechanism that achieves efficient reuse and task-adaptive transfer of multi-expert policies.
    \item We establish a reliable interaction paradigm using VLMs as semantic interfaces, achieving feasible mapping from environmental semantics to physical actions.
\end{itemize}
\begin{figure*}[t] 
    \centering 
    \includegraphics[width=\textwidth]{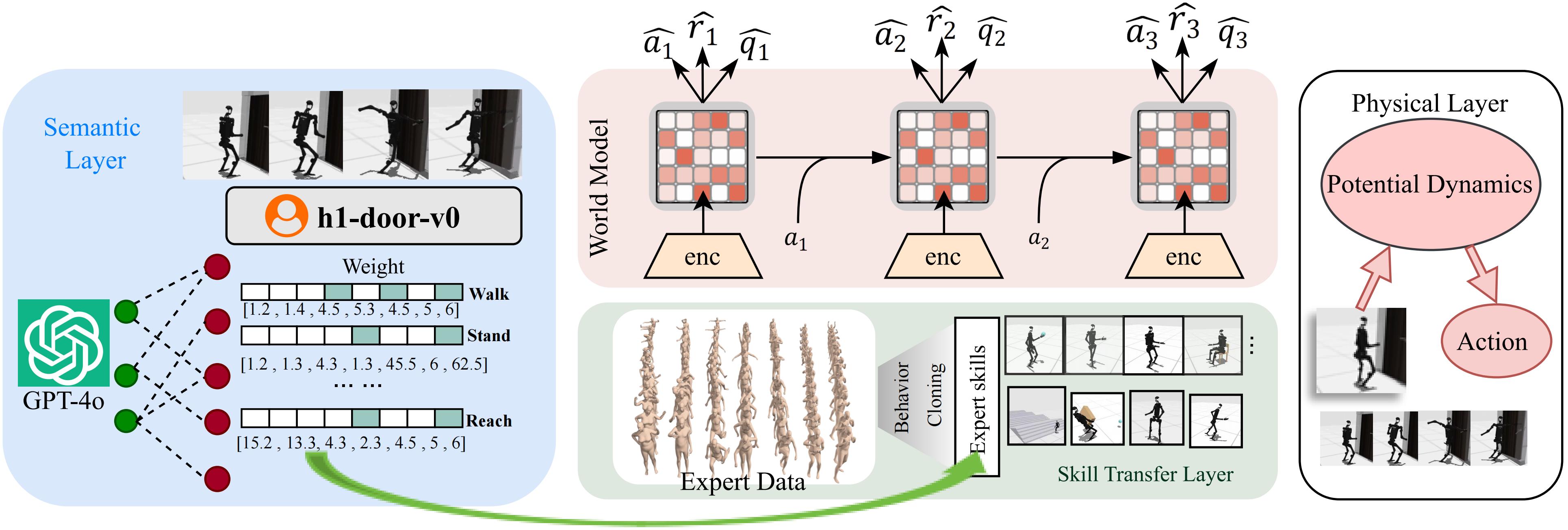} 
    \caption{ This illustrates the three-tier architecture of the \textbf{MetaWorld} framework: the semantic layer parses observation into executable skill sequences via a vision-language model; the skill transfer layer integrates expert policy priors through a hierarchical world model and enables dynamic adaptation; the physical layer performs precise control in a compact state space using a latent dynamics model.} 
    \label{fig:zhanshi} 
\end{figure*}

\section{Related Works}

\subsection{World Models in Robotics}

World models facilitate policy learning by predicting environment state transitions, emerging as a critical tool for enhancing sample efficiency and generalization in robotics~\cite{li2025robotic}. Mainstream approaches, such as the Dreamer series~\cite{hafner2020dreamer,hafner2021dreamerv2,hafner2025dreamerv3,hafner2025dreamer4}, perform policy optimization through latent space prediction, while model-based reinforcement learning methods like TD-MPC2~\cite{hansen2024tdmpc2} further achieve high-performance control in complex dynamic environments. However, most existing world models lack hierarchical depth, hindering the decomposition of high-level semantic tasks into executable physical action sequences, thereby limiting their direct application to long-horizon, multi-step embodied tasks~\cite{fujii2025real}. The hierarchical world model architecture proposed in this paper introduces explicit separation between the semantic planning layer and physical execution layer, enabling the world model to simultaneously handle semantic parsing of language instructions and dynamic prediction of physical environments, effectively bridging the gap between task planning and action generation.

\subsection{Skill Transfer Learning}

Transfer learning facilitates robotic generalization through the systematic reuse of pre-acquired skills. Current robot control approaches include: domain adaptation (e.g., domain randomization) to bridge the sim-to-real gap~\cite{VarianceReducedDomainRandomization}; policy fine-tuning (e.g., progressive networks) for local adjustments~\cite{fehring2025growing}; and meta-learning (e.g., MAML) for fast few-shot adaptation~\cite{FinnAL17,Kayaalp2022Dif-MAML}. However, these methods face two major bottlenecks: a heavy reliance on target-domain data or multi-stage training, both of which limit their practical deployment; and a failure to achieve real-time, millisecond-level adaptation in the face of dynamic disturbances. To address these issues, we propose a dynamic expert selection and motion prior fusion mechanism. By constructing a multi-expert policy library, our approach dynamically selects the most relevant strategies within a Model Predictive Control (MPC) framework to enable real-time policy adjustment. This approach maintains the efficiency of expert policies while significantly enhancing adaptability and generalization in unstructured environments.

\subsection{VLM Applications in Real-time Task Parsing}

Vision-Language Models (VLMs) have demonstrated remarkable proficiency in open-vocabulary perception, scene understanding, and high-level reasoning, and have been widely applied to high-level task planning and semantic guidance in robotics. Representative works such as VIMA and RT-2 attempt to directly use VLMs to generate robot action sequences or serve as front-ends for symbolic planners~\cite{brohan2023rt2,huang2024rekep}. However, these methods commonly face the ``symbol grounding'' problem: VLM-generated plans often ignore the robot's kinematic and dynamic constraints as well as environmental physics, resulting in infeasible plans. In this work, we mitigate the physical limitations of VLMs by restricting their role to high-level semantic parsing and mapping their outputs onto a set of pre-validated, physically feasible expert policies. Concurrently, our framework leverages the semantic understanding and task decomposition strengths of VLMs to process open-ended environmental semantics and achieve dynamic replanning through closed-loop feedback during training, guiding complex tasks to reference fundamental expert policies.

\section{Method}

\subsection{MetaWorld Hierarchical Architecture}
The central idea of \textbf{MetaWorld} is to decompose the robotic control problem into two distinct layers: a semantic planning layer responsible for interpreting task intent, and a physical execution layer responsible for generating physically feasible actions. This hierarchical design can be formalized as:
\begin{equation}
\pi(a_t \mid s_t, \mathcal{T}) = \pi_{\text{phys}}(a_t \mid s_t, \pi_{\text{sem}}(\mathcal{T})),
\end{equation}
where $\pi_{\text{sem}}$ maps the task description $\mathcal{T}$ to a semantic plan, and $\pi_{\text{phys}}$ generates specific actions $a_t$ based on the current state $s_t$ and the semantic plan. This architecture enables independent optimization of semantic understanding and physical control components while maintaining overall optimality. The optimization objective is to maximize the expected cumulative reward $J(\pi) = \mathbb{E}[\sum \gamma^t r(s_t, a_t)]$, where hierarchical optimization effectively addresses the distinct challenges at the semantic and physical levels.

\subsection{Semantic Planning and Symbol Grounding}
The semantic planning layer employs a Vision-Language Model (VLM) to map natural language task descriptions to expert policy weights. Unlike traditional methods, we constrain the VLM output to an expert weight vector $\mathbf{w}$ rather than direct actions:
\begin{equation}
\mathbf{w} = f_{\text{VLM}}(\mathcal{T}, \mathcal{E}),
\end{equation}
the key innovation of this design lies in transforming the symbol grounding problem into a linear combination of expert policies. The VLM generates expert weights through carefully engineered prompts, with the response $R$ processed by a parsing function to obtain normalized weights $w_i = \exp(\text{extract}_i(R)) / \sum_j \exp(\text{extract}_j(R))$. Since each expert policy $\pi_{\text{exp}}^i$ is physically feasible, the generated semantic plan $\pi_{\text{sem}}(\mathcal{T}) = \sum_i w_i \pi_{\text{exp}}^i$ naturally satisfies physical constraints. The symbol grounding error is bounded within the range of expert policy differences, significantly outperforming direct action generation methods.

\subsection{Dynamic Adaptation Mechanism}
To address dynamic environmental changes, we introduce a state-aware expert selection mechanism. Based on the current state $s_t$, we construct a selection probability distribution:
\begin{equation}
p(i \mid s_t) = \frac{\exp(\phi(s_t)^\top \psi(\pi_{\text{exp}}^i))}{\sum_{j=1}^{K} \exp(\phi(s_t)^\top \psi(\pi_{\text{exp}}^j))},
\end{equation}
where $\phi$ is a state encoding function and $\psi$ is an expert feature extraction function. The VLM-generated semantic weights $w_i$ are fused with the dynamic selection probabilities $p(i|s_t)$ to obtain the final weights $\tilde{w}_i(s_t, \mathcal{T}) = \alpha w_i + (1-\alpha) p(i|s_t)$. This fusion mechanism theoretically combines the advantages of long-term task planning and short-term state adaptation. The parameter $\alpha \in [0,1]$ controls the relative importance of semantic planning versus state awareness, enabling a balance between task consistency and environmental adaptability through appropriate adjustment of $\alpha$. The reference expert action $a_{\text{ref}} = \sum_i \tilde{w}_i(s_t, \mathcal{T}) \pi_{\text{exp}}^i(s_t)$ provides a high-quality initial solution for the physical execution layer.

\subsection{Physical Execution and Online Optimization}
The physical execution layer employs the TD-MPC2 algorithm, constructing a latent dynamics model for Model Predictive Control (MPC). Observations $o_t$ are encoded into latent states $z_t = f_{\text{enc}}(o_t)$, and state evolution is predicted through the dynamics model $z_{t+1} = f_{\text{dyn}}(z_t, a_t)$. The MPC optimization problem solves for the optimal action sequence over a future horizon $H$:
\begin{equation}
\mathbf{a}_{t:t+H-1}^* = \arg\max_{\mathbf{a}_{t:t+H-1}} \mathbb{E}\left[ \sum_{k=0}^{H-1} \gamma^k r(z_{t+k}, a_{t+k}) + \gamma^H V(z_{t+H}) \right],
\end{equation}
the expert-guided action $a_{\text{ref}}$ is incorporated into the optimization objective $\mathcal{L}_{\text{total}} = \mathcal{L}_{\text{TD}} + \lambda \|a_t - a_{\text{ref}}\|^2$, where the Temporal-Difference (TD) learning loss $\mathcal{L}_{\text{TD}} = \mathbb{E}[\|Q(z_t, a_t) - (r_t + \gamma Q(z_{t+1}, \pi(z_{t+1})))\|^2]$ ensures accurate value function estimation. This design maintains online adaptation capabilities while leveraging expert knowledge to accelerate the learning process. TD-MPC2 employs quantile regression to learn the value function, achieving policy improvement through minimization of TD errors.

\subsection{Theoretical Analysis and Implementation}
Based on contraction mapping theory, we prove algorithm convergence under appropriate parameter selection. The value iteration operator $\mathcal{T}$ satisfies the contraction property $\|\mathcal{T}Q_1 - \mathcal{T}Q_2\|_\infty \leq \gamma \|Q_1 - Q_2\|_\infty$, guaranteeing convergence of the value function to the optimal solution. Compared to traditional methods, the sample complexity is reduced from $\mathcal{O}(|\mathcal{S}||\mathcal{A}| / [(1-\gamma)^2 \epsilon^2])$ to $\mathcal{O}(1 / [(1-\gamma)^2 \epsilon^2] + K)$, demonstrating the efficiency advantage of knowledge reuse.

\section{Experiment}
We evaluate our method on HumanoidBench~\cite{Sferrazza2024HumanoidBench}, a comprehensive benchmark for humanoid locomotion and manipulation. For locomotion, we select three tasks—walk, stand, and reach—and for manipulation, we include door opening. The locomotion tasks are learned via imitation learning from the AMASS~\cite{AMASS2019ICCV} expert dataset, shaped with trajectory-tracking reward signals. Other skills in the base expert policy repository (e.g., reach) are inherited from TD-MPC2 and are therefore excluded from evaluation. We select two representative state-of-the-art (SOTA) model-based reinforcement learning algorithms as baselines: \textbf{TD-MPC2}, which emphasizes implicit modeling and cross-domain robustness, and \textbf{DreamerV3}, which excels at explicit world model construction and complex visual task planning.
\subsection{Experimental setup}
Our architecture employs the following hyperparameters: the VLM temperature parameter $\tau=0.3$ controls the stochasticity in expert weight generation; the expert guidance weight $\lambda=0.05$ balances TD-learning and expert priors; the MPC planning horizon is set to $H=3$ to ensure real-time performance; the discount factor $\gamma=0.99$ trades off long-term rewards; the learning rate is fixed at $0.001$ to ensure stable convergence; the batch size of $256$ balances training efficiency and memory usage; and the fusion coefficient $\alpha=0.7$ regulates the relative importance of semantic planning and state adaptation. For task success criteria, we follow the definitions provided in HumanoidBench.

\subsection{Validation of Hierarchical Architecture Effectiveness}
\textbf{Purpose of the experiment.}
In this section, we evaluate the effectiveness of our hierarchical architecture against traditional model-based reinforcement learning methods. We first assess the performance of the base expert policy across three locomotion tasks: walk, reach, and stand. Subsequently, on the walk task, we further test the model’s capability to perform complex manipulation tasks after dynamic adaptation.
\begin{figure}[t]
    \centering
    \includegraphics[width=\textwidth]{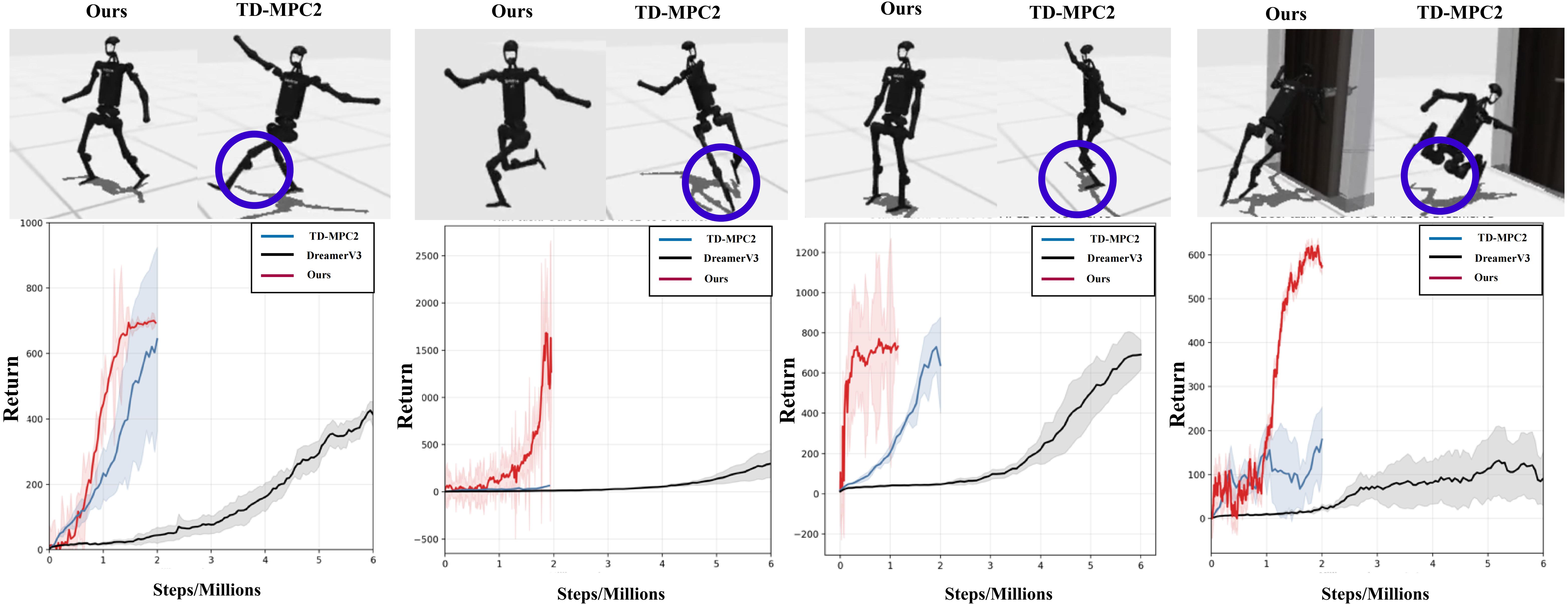}
    \caption{We evaluate our method on locomotion (run, walk, stand) and manipulation (door) tasks against baselines TD-MPC2 and DreamerV3.}
    \label{fig:my_label}
\end{figure}


\begin{table}[t]
\centering
\fontsize{9.5pt}{12pt}\selectfont 
\renewcommand{\arraystretch}{1.5}
\setlength{\tabcolsep}{4pt}

\begin{tabularx}{\textwidth}{l l X X >{\columncolor{blue!10}\centering\arraybackslash}X >{\centering\arraybackslash}X}
\toprule
\textbf{Task} & \textbf{Metric} & \centering\textbf{DreamerV3} & \centering\textbf{TD-MPC2} & \centering\textbf{Ours} & \textbf{Imp. (\%)} \\
\midrule
\multirow{2}{*}{\textbf{Stand}} & Ret. $\uparrow$ ($\pm$ Std) & $699.3 \pm 62.7$ & $749.8 \pm 54.3$ & $\mathbf{793.4 \pm 13.5}$ & 5.8 \\
                                & Conv. $\downarrow$ (M)      & $5.5$            & $1.9$            & $\mathbf{0.8}$            & 57.9 \\
\midrule
\multirow{2}{*}{\textbf{Walk}}  & Ret. $\uparrow$ ($\pm$ Std) & $428.2 \pm 14.5$ & $644.2 \pm 162.3$ & $\mathbf{701.2 \pm 7.6}$  & 8.8 \\
                                & Conv. $\downarrow$ (M)      & $6.0$            & $1.8$            & $\mathbf{1.4}$            & 22.2 \\
\midrule
\multirow{2}{*}{\textbf{Run}}   & Ret. $\uparrow$ ($\pm$ Std) & $298.5 \pm 84.5$ & $66.1 \pm 4.7$   & $\mathbf{1689.9 \pm 13.6}$ & 2456.3 \\
                                & Conv. $\downarrow$ (M)      & $6.0$            & $2.0$            & $\mathbf{1.9}$            & 5.0 \\
\midrule
\multirow{2}{*}{\textbf{Door}}  & Ret. $\uparrow$ ($\pm$ Std) & $165.8 \pm 50.2$ & $179.8 \pm 52.9$ & $\mathbf{680.0 \pm 50.0}$ & 278.3 \\
                                & Conv. $\downarrow$ (M)      & $9.0$            & $2.0$            & $\mathbf{1.7}$            & 15.0 \\
\midrule
\multirow{2}{*}{\textbf{Avg.}}  & Ret. $\uparrow$             & $398.0$          & $410.0$          & $\mathbf{966.1}$          & 135.6 \\
                                & Conv. $\downarrow$ (M)      & $6.6$            & $1.9$            & $\mathbf{1.5}$            & 24.9 \\
\bottomrule
\end{tabularx}

\vspace{2mm}
\caption{Performance comparison on locomotion and manipulation tasks. All values are reported to one decimal place. The \textbf{Ours} column is highlighted in light blue, and the rightmost column shows the percentage improvement.}
\label{tab:results_adaptive_blue}
\end{table}

\textbf{Result.}
Based on the experimental results shown in Figure~\ref{fig:my_label} and Table~\ref{tab:results_with_improvement}, the \textbf{MetaWorld} framework's comprehensive advantages across four tasks (average return improvement of 135.6\%) demonstrate its hierarchical architecture's effectiveness. For basic locomotion tasks (Stand, Walk, Run), the exceptional performance (especially Run's 2456.3\% improvement) benefits primarily from high-quality motion priors provided by the imitation-learned expert library, which supplies the VLM planner with a physically feasible action space. For the complex Door task, the 278.3\% gain validates the hierarchical innovation: the VLM decomposes the "open door" instruction into sub-actions, while the dynamic selection mechanism composes corresponding primitive experts for execution. The key insight is that complex tasks are achieved through semantic parsing and expert composition rather than end-to-end training. This strategy effectively resolves the symbol grounding dilemma while simultaneously overcoming the sample efficiency bottleneck inherent in end-to-end RL paradigms, offering a novel paradigm for intelligent robot control.

\subsection{Ablation study and stability analysis}
\textbf{Purpose of the experiment.}
To rigorously quantify the contribution of each module within the MetaWorld framework, we perform a systematic series of ablation experiments. First, we ablate semantic planning by uniformly weighting all tasks in the base skill library. Next, we ablate dynamic expert selection by fixing $\alpha=1.0$ during training, thereby disabling state-aware expert selection. Finally, we ablate skill-expert guidance by setting $\lambda=0$ in the TD-MPC2 optimization objective, removing expert action guidance. We present experimental observations and analysis for each ablation.

\begin{figure*}[h] 
    \centering 
    \includegraphics[width=0.9\textwidth]{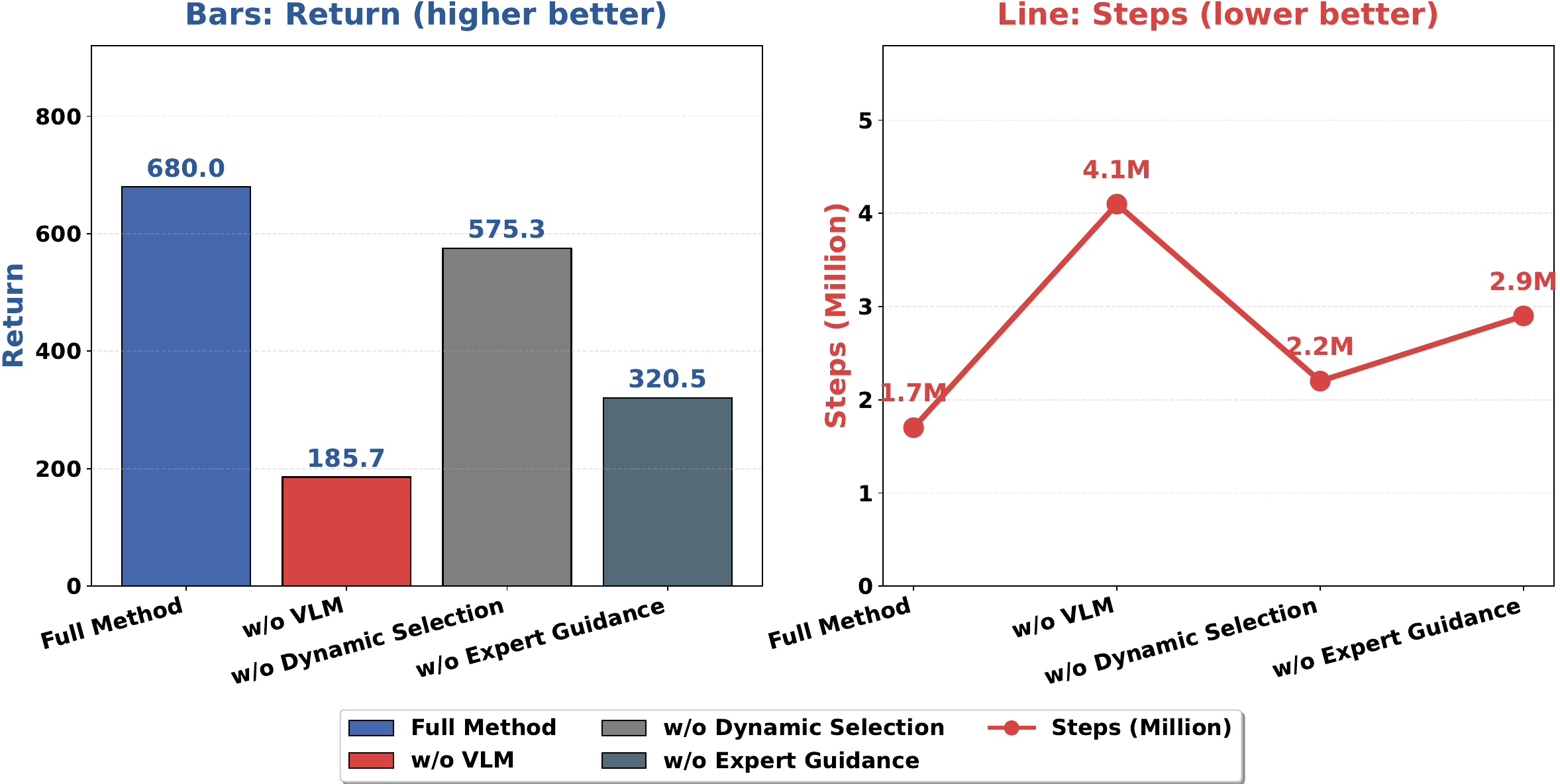} 
    \caption{Ablation study results on the Door task (maximum reward and convergence steps).} 
    \label{fig:ablation} 
\end{figure*}

\textbf{Result.}
As shown in Figure~\ref {fig:ablation}, ablation studies on the Door task reveal how the \textbf{MetaWorld} framework bridges the semantic-physical gap. Removing the Vision-Language Model (VLM) causes a 72.7\% performance collapse, indicating a failure in semantic planning, the robot cannot decompose "open the door" into executable skills like "approach handle–rotate–push/pull." This highlights traditional RL’s inability to solve the symbol grounding problem. Ablating dynamic expert selection results in only a 15.4\% performance drop, demonstrating the framework’s robustness: without online adaptation, the pre-trained expert base still provides basic motion priors. To rigorously quantify the contribution of each module within the MetaWorld framework, we perform a systematic series of ablation experiments. Removing expert guidance results in a 52.9\% performance loss, highlighting the synergy between imitation learning and model-based RL. Expert policies provide feasible action bounds for online fine-tuning, with both components being essential.

\section{Conclusion}
This study presents \textbf{MetaWorld}, a hierarchical world model framework that bridges the semantic-physical gap in humanoid robot control via VLM-based semantic planning, dynamic expert policy transfer, and latent physical control. On Humanoid-Bench tasks, it achieves an average reward improvement of 135.6\%, substantially enhancing task completion efficiency and motion coherence. While the framework includes a basic imitation learning module to build an expert policy repository, it currently depends on simple trajectory-matching reward mechanisms, which cannot dynamically weigh fine-grained aspects like joint-level motion errors. Furthermore, expert policy selection uses static weighted fusion rather than an intelligent routing mechanism under a Mixture-of-Experts (MoE) architecture, limiting precise skill composition and semantically conditioned switching. The system also lacks a few-shot generalization ability, shows constrained adaptability and scalability with novel complex task combinations, and does not explicitly address gradient interference from multi-skill coordination. Future work will iteratively upgrade the framework through dynamic reward shaping for imitation learning, an MoE-based semantic-aware routing mechanism, and enhanced few-shot transfer generalization, advancing toward a more robust and adaptive framework for complex control and manipulation demands.

\bibliography{iclr2021_conference}
\bibliographystyle{iclr2021_conference}

\end{document}